\documentclass[11pt,a4paper]{article}
\usepackage[utf8]{inputenc}
\usepackage[english]{babel}
\usepackage{amsmath}
\usepackage{amsfonts}
\usepackage{amssymb}
\usepackage{graphicx}
\usepackage[left=2.54cm,right=2.54cm,top=2.54cm,bottom=2.54cm]{geometry}
\usepackage[numbers]{natbib}
\usepackage{float}
\usepackage{enumitem}
\usepackage{appendix}
\title{Locomotion Design for an Internally Actuated Cubic Robot for Exploration of Low Gravity Bodies in the Solar System}
\author{{\small \begin{tabular}{ccc} Alvaro Bátrez & Gustavo Rodriguez-Gomez & Angélica Muñoz-Meléndez\\ Universidad Politécnica de Pachuca & Instituto Nacional de & Instituto Nacional de\\ alvaro\_batrez@upp.edu.mx & Astrofísica, Óptica & Astrofísica, Óptica \\ & y Electrónica & y Electrónica\\ & grodrig@inaoep.mx & munoz@inaoep.mx \end{tabular}}}
\date{}       

\begin{document}

\maketitle

\begin{abstract}
The exploration of asteroids and comets is important in the quest for the formation of the Solar System and it is an important step for human space travel. Moving on the surface of asteroids is challenging for future robotic explorers due to the weak gravity force. In this research, an approach that is based on a new kind of jumping rovers is presented. This robotic platform has internal masses and by spinning up these flywheels and suddenly stopping them, it is feasible to perform a hop from a few meters up to hundreds of meters. In contrast to related works where robotic explorers usually stop a flywheel instantaneously, the INAHOPPER prototype takes advantage of on stopping a flywheel by voltage inversion in a short lapse to modify the launch angle, a very useful action over terrains with different degrees of inclination. This article discusses, first, the dynamics of the rover for a 2D model, second, the control algorithm executed in the prototype for braking the flywheel, and third, the analysis of the performance of the flywheel to make simulations of the trajectories over an asteroid.
\end{abstract}

\section{Introduction}

The exploration of small primitive bodies such as asteroids, comets, and small moons, is a priority for the understanding of the formation of the Solar System, the origin of life, and it is an important step to the journey to Mars and beyond \cite{national2011vision}. The milli-micro gravity force of these small bodies is a constraint for the traditionally wheeled rovers. The lack of traction causes that the maximum speed that can be reached is about 1.5 $mm/s$ \cite{wilcox2000muses}. Bouncing and dust are other important inconveniences for wheeled rovers. For these reasons, it is convenient to analyze alternative motion strategies that take advantage of low gravity.

Various solutions for mobile surface platforms have been proposed such as rocket thruster, legged and jumping robots. Each one of the mentioned ways of motion presents advantages and disadvantages, and have different purposes \cite{nasa2015nasa}. Mobility by rocket thrusters would land few times to collect soil samples and data from the on-board instruments. A drawback of this alternative is the limited number of locations that the spacecraft might be able to visit \cite{nasa2008chopper}, \cite{castillo2012expected}. Mobility based on legs is mechanically complex because it requires an anchoring system and also is slower than the other mentioned solutions. However, a legged robot could reach some unattainable places for mobile robots that do not rely on legs, such as cliffs \cite{helmick2014small}, \cite{parness2017lemur}. In the present research we focus on mobility by jumping because it has the advantage that using a simple mechanism jumping robots are capable of making long traverses, and also, since the actuation is internally encapsulated there is no risk of stuck by dust \cite{castillo2012expected}, \cite{herrmann2011mobility}, \cite{hockman2017design}. The INAHOPPER has a minimalistic design with low weight, small dimensions, and low cost. Its inertial wheels are held within the explorer, the movement is produced by slowly accelerating these wheels and suddenly stopping them, generating in that way a braking torque and the consequent transfer of energy from the wheels to the chassis that causes the robot jumps \cite{herrmann2011mobility}, \cite{dietze2010landing}, \cite{allen2013internally}. The trajectory of the hop is defined by the shape of the robot and the surface inclination. \citet{hockman2017design} have established that the best morphology of a \textit{hopper} is a cubic shape since it always offers a launch angle of 45º over flat terrains.

In recent related works, the \textit{hopper} robot has an instantaneous mechanical brake that is applied by both, a rubber band \cite{hockman2017design} and an impact hammer \cite{hockman2015design}. In contrast, in this research, the application of controlled braking of the wheels by voltage inversion over surfaces with different degrees of inclination, with the aim of changing the launch angle for always getting the desired motion regardless the slope of the ground on which is located the robot is proposed. This choice aims at protecting eventual damages that might happen to the mechanical components of the robot for a sudden uncontrolled brake. As a first approximation, a 2D ballistic flight and one single flywheel is considered and also it is assumed that the center of mass of the INAHOPPER is on its geometric center.

In the next rest of this section significant research about hopping space robots is briefly revised. The first is a robot called MASCOT \cite{herrmann2011mobility}, \cite{dietze2010landing}. It has been presented by the European Space Agency and the German Aerospace Center. This robot is currently on the asteroid 1993 JU3, on-board the spaceship HAYABUSA-2. It uses a mobility strategy that is similar to the proposed in this research; however, MASCOT only needs one single mass, that is outside its gravity center, which enables the explorer jumping to a different location on the surface of an asteroid. MASCOT is unable to focus on a particular place of interest since it has been designed to collect information from random positions.

\citet{kato2017distance} have presented a small rocket-propelled robot, that also has an inertial load within its interior. This work uses a small main rocket engine to lift off from the ground and a second engine to stabilize the traverse trajectory. When the robot is flying, a reaction wheel starts to rotate for adjusting the thrusting direction of both engines. This strategy needs one shot and it can be very challenging to control under low gravity.

The research of \citet{hockman2017design} is based on the work of \citet{muehlebach2017nonlinear} and is adapted to a situation of low gravity. The validation of the experiments and the response of the algorithms is conducted using a crane type system. This work leaves several open problems, such as the development of realistic contact models of sandy terrain (regolith) found in asteroids; the improvement of the reliability of the maneuvers on rocky terrains, and the development of Simultaneous Localization and Mapping (SLAM) techniques to know the complexity of the environment, among others. The work of \citet{hockman2017design} is the main reference of our own research.

The main contribution has been to control the inertia of the flywheels in such a way that the launch angle of the jump can be deflected effectively in order to reach a given location regardless the surface inclination on which the robot is located. It is worth to mention that in related works, the braking of the flywheel is instantaneous thus giving always the same launch angle.

The rest of this document is organized as follows, Section II reviews the dynamic of the robot under very low gravity, also the boundaries and assumptions of the investigation are established. Section III proposes a solution to perform effective mobility of the cubic robot. Section IV describes the experiments and numerical simulations to validate the proposed solution that were conducted. Section V discusses the results of the experiments, and finally, Section VI summarizes the results and closes with ideas for future work.

\section{Modeling the Dynamics of a Cubic Robot}

\begin{figure}[H]\centering
    \includegraphics[width=0.5\textwidth]{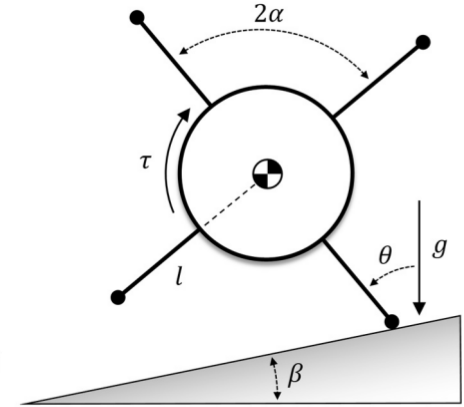}
    \caption{Robot model in 2D \cite{hockman2017design}.}
    \label{figHopperModelo2D}
\end{figure}

The representation of the robot for a 2D model and just one flywheel is illustrated in Figure \ref{figHopperModelo2D}. The explorer is represented as a circle with four rigid bars attached to it. At the center of mass, there is a DC motor with an inertial load that provides a torque to the robot.

\begin{table}[H]\centering
    \begin{tabular}{clc}
        \textbf{Symbol} & \textbf{Description} & \textbf{Units}\\
        \hline
        $\theta$ & Hopper's angle & \textit{degrees}\\
        $\beta$ & Surface slope & \textit{degrees}\\
        $l$ & Spike's length & $m$\\
        $I_f$ & Flywheel's rotational inertia & $kg \cdot m^2$\\
        $I_p$ & Platform's rotational inertia & $kg \cdot m^2$\\
        $\tau$ & Flywheel's torque & $N \cdot m$\\
        $2 \alpha$ & Angle between spikes & \textit{degrees}\\
        $\omega_f$ & Flywheel angular speed & $rad/s$\\
        $g$ & Gravity acceleration & $m/s^2$\\
        $m_p$ & Platform's mass & $kg$
    \end{tabular}
    \caption{Parameters of the 2D model \cite{hockman2017design}.}\label{table1}
\end{table}

To realize a jump maneuver, a reaction wheel is slowly accelerated and suddenly stopped to generate the momentum that will provide a ballistic trajectory. A hop maneuver is divided into two phases:

\begin{itemize}
\item \textit{Leverage phase}. The time elapsed between the moment when the robot is at a rest position and the time when any of the spikes have no contact with the ground.
\item \textit{``Fly'' phase}. The time elapsed between the moment when the platform leaves the ground and the moment when it falls again.
\end{itemize}

\begin{figure}[ht]\centering
\includegraphics[width=1\textwidth]{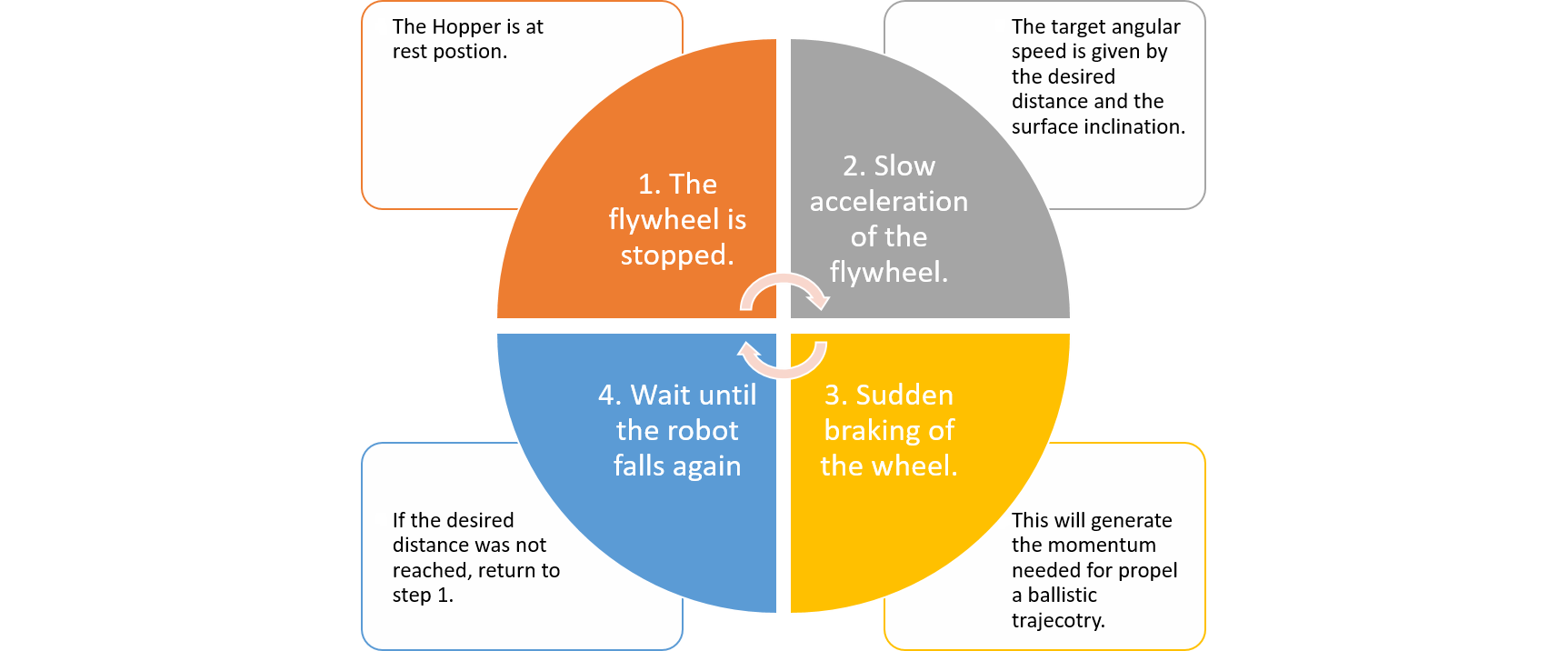}
\caption{Diagram of a complete jump maneuver.}
\label{figureCircle}
\end{figure}

Below it is explained how a complete maneuver of the robot is made to perform a jump to the target area. Figure \ref{figureCircle} illustrates the process.

Considering the objective distance $d_{obj}$ and the surface inclination $\beta$, first, the flywheel is slowly spun up with motor torque $\tau < \tau_{min}$, Equation \eqref{hop3}, to prevent the robot loses contact with the ground in advance. The target angular speed $\omega_f$ is given by $d_{obj}$ and $\beta$, Section \ref{speed-wf}. When $\omega_f$ is reached, the second step is to stop the  motor applying a voltage inversion brake at a specific time. This time is also related to $d_{obj}$ and $\beta$. Consequently, a ballistic trajectory is initiated and the process must wait until the platform falls again. As a final step, it is necessary to verify if $d_{obj}$ was met, if so, returns success. Otherwise, a repetition of the complete process to perform a new jump would be needed.

Table \ref{table1} describes the parameters of the \textit{hopper} for a 2D model.

\subsection{Hopping Dynamics and Instant Braking of the Flywheel}

We assume that there are not slips on the edge of the spikes. The slips and bounces are governed by the friction with the regolith and rocks on the surface of the comets or asteroids. These interactions are beyond the scope of current research.

Equations \eqref{hop8} -- \eqref{hop10} are the main expressions for relating the electrical brake of the flywheel and the hopping trajectory of the robot. For a deeper analyze of this equations please consult Appendix \ref{app1}.

The hop angle $\theta_h$ is obtained by the geometry of the robot $\alpha$, and the slope of the surface $\beta$. For horizontal terrains $\beta = 0$º, this angle is always 45º, Equation \eqref{hop8}
\begin{equation}\label{hop8}
\theta_h = \alpha + \beta.
\end{equation}

This is the primary reason because the hopper is a cube instead other polyhedron.

The horizontal distance is given by a simple formula of projectile movement, indicated in Equation \eqref{hop9}
\begin{equation}\label{hop9}
d_h = \frac{v_h^2 \sin (2 \theta_h)}{g}.
\end{equation}

From Equations \eqref{hop4} -- \eqref{hop7}, \eqref{hop8}, \eqref{hop9} it is possible to obtain the flywheel speed required to cover a horizontal distance, , Equation \eqref{hop10}
\begin{equation}\label{hop10}
\omega_f = \sqrt{\frac{d_h g}{\eta^2 l^2 \sin ( 2 (\alpha + \beta ) ) }}.
\end{equation}

\section{Proposed Solution}

In this section, a solution for controlling the jumps of INAHOPPER cube is presented in detail. First, as the surface of an asteroid or comet is rarely flat, is a motivation of this research to deflect the launch angle depending on the slope of the ground in order to maximize the hop distance; for this, it will be necessary to extend the levering phase. Second, as a sudden braking by impact of the inertial wheel might be harmful for the robot, we propose to stop the flywheel by voltage inversion. This consideration contributes also to decrease the weight of the robot.

\subsection{Controlled Braking of the Flywheel}

\citet{hockman2017design} have established that since the momentum transfer is no longer instantaneous, it is considered that the braking torque $\overline{\tau}$ and the angular momentum of the flywheel are related by Equation \eqref{hop11} where $\Delta t$ is the braking time
\begin{equation}\label{hop11}
\overline{\tau} \Delta t = I_f \omega_f.
\end{equation}

For big jumps we can assume that $\overline{\tau} \gg m_p g l \sin \theta$, so Equation \eqref{hop2} can be approximated by Equation \eqref{hop12}
\begin{equation}\label{hop12}
\ddot{\theta} \approx \frac{-\tau}{I_p + m_p l^2}.
\end{equation}

The new expressions will be: the launch velocity $v_h$ remains equal as described in Equation \eqref{hop7}; the horizontal distance $d_h$ is also given by the projectile formula as in Equation \eqref{hop9}. The launch angle $\theta_h$ can be determined by \cite{hockman2017design}
\begin{equation}\label{hop13}
\theta_h = \alpha + \beta - \frac{\eta I_f \omega_f^2}{2 \overline{\tau}}.
\end{equation}

By a controlled braking torque $\overline{\tau}$ of the flywheel, the launch angle can be deflected. The goal is to reach a launch angle of 45º for maximizing the jump distance as expressed by Equation \eqref{hop13}.

\subsection{Braking Torque, Launch Angle and Jump Distance Relationships}

To better understand the implementation of a controlled braking of the inertial wheel, some numerical simulations illustrating the relationships among the braking torque $\overline{\tau}$, the launch angle $\theta_h$, and the jump distance $d_h$, have been generated, as described below.

Taking into account an angular speed of the flywheel of $\omega_f = 3500\ rpm$ or $366.5\ rad/s$, a surface slope of $\beta = 15$º, and the physical characteristics of the robot summarized in Table \ref{tableParameters}. The objective of these simulations is to know how $\theta_h$ and $d_h$ change in function of $\overline{\tau}$. For these examples $\overline{\tau}$ ranges from $10^{-2}\ Nm$ to $10^{1}\ Nm$. It is expected that the maximum jump distance and a launch angle of 45º can be found at the same magnitude of the braking torque.

\begin{figure}[ht]\centering
\includegraphics[width=0.6\textwidth]{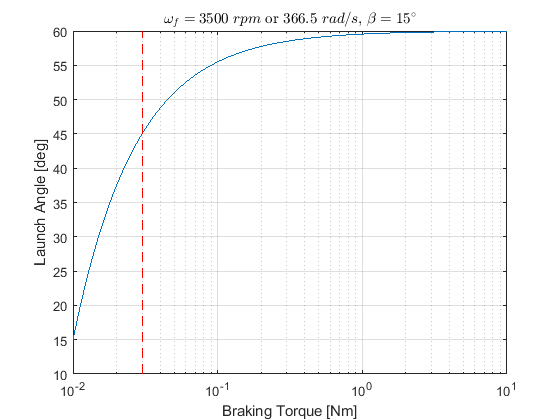}
\caption{Resulting $\theta_h$ as function of $\overline{\tau}$.}
\label{figTauTheta}
\end{figure}

\begin{figure}[ht]\centering
\includegraphics[width=0.6\textwidth]{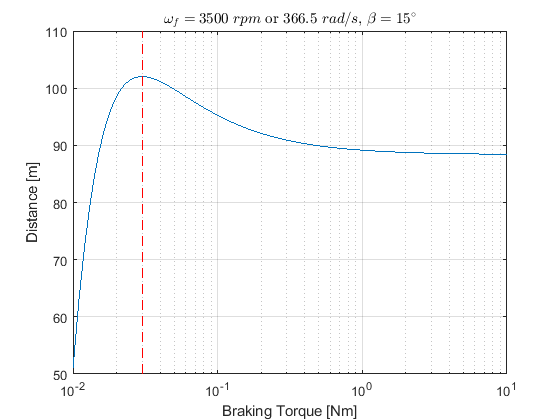}
\caption{Resulting $d_h$ as function of $\overline{\tau}$.}
\label{figTauDh}
\end{figure}

Figure \ref{figTauTheta} shows the obtained results of this test about how the launch angle changes, while the braking torque increases. In the same way, Figure \ref{figTauDh} illustrates the relationship between the jump distance and the braking torque. Both images demonstrate that the application of a torque of $\overline{\tau} = 0.03\ Nm$ (red dotted line) under a gravity acceleration of $77\ \mu m / s^2$ enables the reach of a maximum distance about $d_h = 102.03\ m$ and a launch angle of $\theta_h = 45$º.

\subsection{Adjust the Trajectory for Uneven Terrains}\label{speed-wf}

The angular speed $\omega_f$ cannot be calculated directly, as we did in Equation \eqref{hop10}. It is necessary to obtain it numerically varying values of target angular speed, the braking time in function of the distance and the surface inclination. Applying Equations \eqref{hop3}, \eqref{hop7}, \eqref{hop9}, \eqref{hop11} and \eqref{hop13} it is feasible to generate such variations, as well as other characteristics such as the minimum speed-up time of the flywheel, the launch speed and the ``fly time''.

\subsection{Design of the Controller}

The next step is to develop an angular speed controller that enables the robot to brake the flywheel from $\omega_f$ to 0, in a time $\Delta t$. Stop the flywheel in a short lapse can be challenging for a small robot, whose power consumption and constrained dimensions are factors to considered. In this work, we implemented a state feedback control with pole placement, based on continuous time for the simulations and on discrete time for real experiments.

\subsubsection{DC Motor Model}

As we see previously, the actuator of the hopper consists of an inertial load or flywheel attached to a DC motor. For developing a state feedback control it is essential to get the model of the motor. This model is common in the field literature and can be found in \cite{messner2017dc}. Taking the angular speed $\omega_f(t)$ and the circuit current $\imath(t)$ as the output state variables, the state-space representation is given by Equation \eqref{dC10}
\begin{eqnarray}\label{dC10}
\dot{\mathbf{x}} = \begin{bmatrix} -\dfrac{b}{J} & \dfrac{K}{J}\\ -\dfrac{K}{L} & -\dfrac{R}{L} \end{bmatrix} \mathbf{x} + \begin{bmatrix} 0\\ \dfrac{1}{L} \end{bmatrix} v, & y = \begin{bmatrix} 1 & 0 \end{bmatrix}\mathbf{x},
\end{eqnarray}
where the state vector $\mathbf{x} = \begin{bmatrix} \omega_f(t) & \imath(t) \end{bmatrix}^T$, $\mathbf{\dot{x}} = d\mathbf{x}/dt$, $R$ is the armature resistance, $L$ is the electric inductance, $K$ is both the electromotive force (emf) and the motor torque, $J$ is the inertial load on the rotor, $b$ is the motor viscous friction, $v$ is the input voltage.

Applying the Laplace Transform to Equation \eqref{dC10} the transfer function is represented as in Equation \eqref{dC5}
\begin{equation}\label{dC5}
\frac{\omega_f(s)}{V(s)} = \frac{K}{(R + L s)(J s + b) + K^2},
\end{equation}
where the output angular speed $\omega_f(s)$ can be controlled through the input voltage $V(s)$.

\subsubsection{State Feedback by Pole Placement}\label{polePlace}

For continuous linear time invariant systems, the state space representation is straightforward and can be found in \cite{ogata2010ingenieria}. For tracking a reference $r(t)$ \cite{sename2017pole}, in this case, the desired angular speed $\omega_f(t)$ of the inertial wheel,  the control signal, Equation \eqref{dC15}, can be selected as follows
\begin{equation}\label{dC15} 
u(t) = -\mathbf{K x}(t) + \mathbf{G}r(t),
\end{equation}
where \textbf{K} is the state-feedback gain matrix.

For discrete linear time-invariant systems, required to be implemented using a microcontroller, the state space representation is also straightforward \cite{fernandez2014sistemas}. The control signal for tracking $r(k)$ is given by Equation \eqref{dC19}
\begin{equation}\label{dC19}
u(k) = -\mathbf{K x}(k) + \mathbf{G}r(k).
\end{equation}

\textbf{G} is the extra gain needed to compensate the steady state error. For a continuous time is given by Equation \eqref{dC20}
\begin{equation}\label{dC20}
\mathbf{G} = (\mathbf{C}(-\mathbf{A}+\mathbf{BK})^{-1} \mathbf{B})^{-1},
\end{equation}
and for discrete time by Equation \eqref{dC21},
\begin{equation}\label{dC21}
\mathbf{G} = (\mathbf{H}(\mathbf{I}-\mathbf{\Phi}+\mathbf{\Gamma K})^{-1} \mathbf{\Gamma})^{-1},
\end{equation}
where, in this case, $\mathbf{I}$ is a $2 \times 2$ identity matrix. For both, $\mathbf{K}$ is a $1 \times 2$ matrix that can be found by Ackermann's formula \cite{ogata2010ingenieria}. The matrices $\mathbf{\Phi}$, $\mathbf{\Gamma}$ and $\mathbf{H}$, are the discrete equivalents of $\mathbf{A}$, $\mathbf{B}$ and $\mathbf{C}$, respectively, and have the same size.

\section{Implementation}

\subsection{Deflecting the Launch Angle of the INAHOPPER}\label{deflecting}

To validate the performance of the \textit{hopper} presented in Section III the implementation was divided into two phases:
\begin{enumerate}
    \item \textbf{The viability of an effective, fast motor braking by voltage inversion.} In this part, real experimentation has been carried out through the INAHOPPER prototype using a DC motor, speed sensor, energy supply, motor driver and a microcontroller, as indicated in Section \ref{physParams}. 
    \item \textbf{The parabolic flight of the robot.} The second phase checks the numerical simulations of the trajectory of the \textit{hopper} under the gravity of the asteroid Itokawa. These simulations changes depending on the stopping performance of the flywheel obtained in the first phase.
\end{enumerate}

The first step for performing a jump to a desired distance $d_h$ is getting the angular speed of the flywheel $\omega_f$ and the braking time $\Delta t$ to deflect the launch angle of the robot $\theta_h$, with the aim of maximizing the jump distance of the robot despite the inclination of the ground. The parameters required to find $\omega_f$ and $\Delta t$ are shown in Table \ref{tableParameters}.

The model of the \textit{hopper} is a $10 \times 10 \times 10\ cm$ cube with an inertial wheel of aluminum of a diameter of $6\ cm$ and $76\ g$ of weight. Figure \ref{figProto} gives an approach of the model.  The total weight of the robot is about $1.5\ kg$ since current 1U cubesats are around that size \cite{woellert2011cubesats}.

\begin{table}[ht]\centering
\begin{tabular}{clc}
\textbf{Symbol} & \textbf{Description} & \textbf{Value}\\
\hline
$g$ & Itokawa's mean gravity acceleration & 77 $\mu m/s^2$\\
$\beta$ & Surface inclination & -30º$< \beta <$30º\\
$\alpha$ & Half the angle between spikes & 45º\\
$l$ & Spike's length & 0.071 $m$\\
$m_p$ & \textit{Hopper}'s mass & 1.5 $kg$\\
$m_f$ & Flywheel's mass & 0.076 $kg$\\
$I_p$ & \textit{Hopper}'s rotational inertia & 25e-4 $kg m^2$\\
$I_f$ & Flywheel's rotational inertia & 3.42e-5 $kg m^2$
\end{tabular}
\caption{Parameters required to calculate the ballistic trajectory.}\label{tableParameters}
\end{table}

Apart from the aforementioned parameters $d_h$, $\omega_f$, $\Delta t$, and $\theta_h$; other parameters are also important to know the ideal trajectory of a jump. The speed-up time $T_{min}$ is the minimum time to reach the objective angular speed of the flywheel from rest position. $L_s$ is the velocity at which the \textit{hopper} will be fired or launch speed. $T_f$ is the time the robot takes to reach its destination or ``fly'' time.

\begin{table}[ht]\centering
\begin{tabular}{llrrrrrr}
\textbf{Distance} & $\mathbf{d_h}\textbf{[m]}$ & \textbf{5} & \textbf{10} & \textbf{30} & \textbf{50} & \textbf{70} & \textbf{100}\\
\hline
\textbf{Angular speed} & $\omega_f$[rad/s] & 81.3 & 115.0 & 198.9 & 256.7 & 303.7 & 363.0\\
\textbf{Speed-up time} & $T_{min}$[s] & 375.0 & 531.0 & 919.0 & 1186.0 & 1403.0 & 1677.0\\
\textbf{Braking time} & $\Delta t$[s] & 2.78 & 1.97 & 0.97 & 0.80 & 0.69 & 0.56\\
\textbf{Launch speed} & $L_s$[cm/s] & 2.0 & 2.8 & 4.8 & 6.2 & 7.3 & 8.8\\
\textbf{Launch angle} & $\theta_h$[degrees] & 43.0 & 43.0 & 46.0 & 45.0 & 44.0 & 45.0\\
\textbf{Fly time} & $T_f$[s] & 347.0 & 491.0 & 900.0 & 1139.0 & 1335.0 & 1610.0
\end{tabular}
\caption{Surface slope $\beta = 20$º.}\label{tab5to100-1}
\end{table}

Table \ref{tab5to100-1} summarizes the calculations for the jump parameters for, respectively, a surface slope of 20º. This Table shows representative distances from $5\ m$ to $100\ m$. It is worth to remark that for negative angles of inclination the braking time is about $\Delta t \approx 0 \ s$.

As it can be seen in Table \ref{tab5to100-1}, the longer the distance $d_h$, the faster the launch speed $L_s$. Reaching distances greater than 100 $m$ is feasible, however, applying the escape velocity might be dangerous. For Itokawa, the escape velocity is about 11.28 $cm/s$ \cite{jpl2003itokawa}. For example, in Table \ref{tab5to100-1} the maximum launch speed $L_s$ is $8.8\ cm/s$.

\subsection{Finding the Physical Parameters of the Flywheel}\label{physParams}

\begin{figure}[ht]\centering
\includegraphics[width=1.0\textwidth]{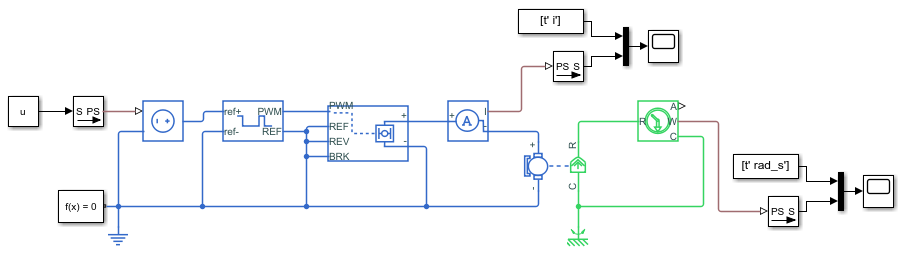}
\caption{Model of a direct-current motor in Simscape\texttrademark.}
\label{figMotorSimscape}
\end{figure}

Simulink Simscape Electronics\textsuperscript{TM} R2017b was used to to obtain the values \textit{b, R, K} and depicted in Table \ref{tabEstimatedParams}; \textit{J} was estimated by the weight and size of the flywheel. Figure \ref{figMotorSimscape} presents the model of a DC motor considering the inherent electrical (resistance, inductance, electro-motive force) and mechanical (shaft inertia, friction) characteristics of the motor; also, the model contains an H-Bridge, a PWM voltage source, a current sensor and a rotational motion sensor.

\begin{figure}[H]\centering
\includegraphics[width=0.5\textwidth]{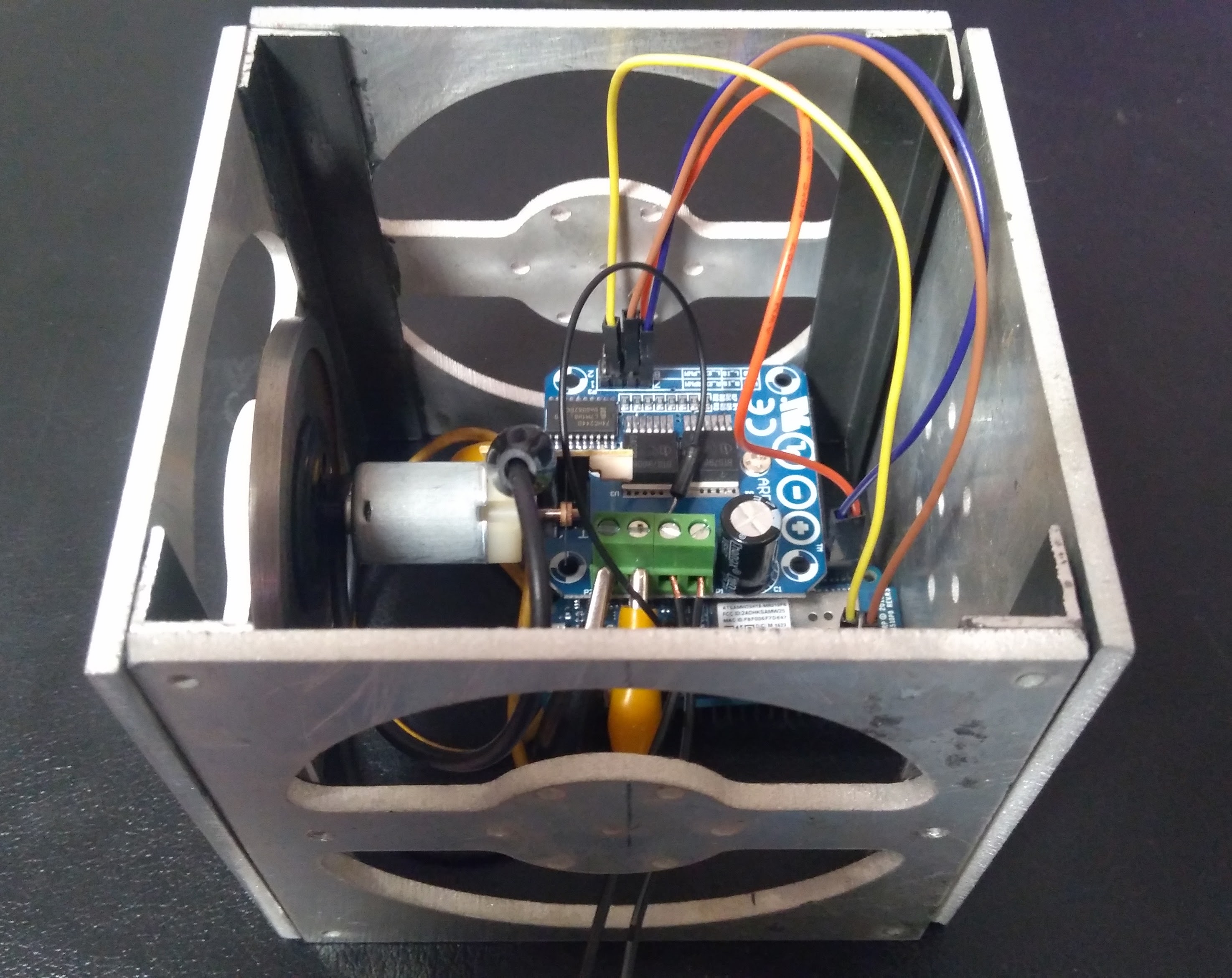}
\caption{Prototype used for the experiments.}
\label{figProto}
\end{figure}

Using measured data from a real system is feasible to get the parameters of the motor. The experiments were conducted using an Arduino board MKR1000 \cite{arduino2018mkr1000}, an H-Bridge BTS7960B \cite{infineon2004bts7960}, a current sensor ACS712 \cite{allegro2007acs712} and a DC motor with rotary optical encoder. Figure \ref{figProto} shows the components of the cubic robot. A DC power supply of 24 V --- 2 A was used.

The obtained values of the electrical and mechanical parameters of the DC motor are presented in Table \ref{tabEstimatedParams}.

\begin{table}[H]\centering
\begin{tabular}{clrc}
\bf{Symbol} & \bf{Description} & \bf{Value} & \bf{Units}\\
\hline
$J$ & Moment of inertia of the flywheel & 3.42e-5 & $kg\ m^2$\\
$b$ & Friction coefficient of the rotor & 2.20e-5 & $N\ m\ s$\\
$K$ & Back electro-motive force constant & 47.96e-3 & $\frac{V}{rad/s}$\\
$L$ & Motor winding inductance & 7.75e-3 & $H$\\
$R$ & Motor winding resistance & 11.36 & $\Omega$
\end{tabular}
\caption{Estimated parameters of the motor.}
\label{tabEstimatedParams}
\end{table}

\subsection{DC Motor Braking by Voltage Inversion}

The parametrized discrete state-space representation with a sample time $T_s = 0.0005\ s$ rounded by 4 decimals is expressed in Equations \eqref{canonD1} and \eqref{canonD2}
\begin{equation}\label{canonD1}
\mathbf{x}(k+1) = \begin{bmatrix} 0.9992 & 0.5821\\ -0.0012 & 0.7242 \end{bmatrix} \mathbf{x}(k) + \begin{bmatrix} 0.0090\\ 0.0252 \end{bmatrix} v(k),
\end{equation}

\begin{equation}\label{canonD2}
y(k) = \begin{bmatrix} 1 & 0 \end{bmatrix}\mathbf{x}(k).
\end{equation}

The main objective is tracking a reference angular speed as faster as it can be with almost no overshoot, in order to prevent that the motor turns backward. To achieve this goal an overshoot $OS\approx0$\% and a settling time $St=0.1$ s were selected. From Section \ref{polePlace} it is possible to compute \textbf{K} and \textbf{G} needed to accomplish the proposed $OS$ and $St$. So the gain matrix $\mathbf{K}$, Equation \eqref{acker} has the values
\begin{equation}\label{acker}
\mathbf{K} = \begin{bmatrix} -0.0246 & -9.4392 \end{bmatrix},
\end{equation}
and the constant $\mathbf{G} = 0.0229$.

\section{Results}\label{secResults}

\subsection{Overview of some speed control responses for a wide variety of target distances and surface inclinations}

The objective of the tests presented in this section is checking the effectiveness of the speed control of the motor by voltage inversion under several angular speeds and braking times. It is expected that the real responses (black lines) follow the model responses (red lines). Table \ref{tabTests} summaries the objective parameters.

\begin{table}[H]\centering
\begin{tabular}{crrrr}
\textbf{Test} & $\omega_f\ [rad/s]$ & $\beta\ [degrees]$ & $\Delta t\ [s]$ & $d_h\ [m]$\\
\hline
2 & 81.16 & 0 & 0.00 & 5\\
3 & 114.88 & 10 & 0.98 & 10\\
4 & 198.76 & 5 & 0.26 & 30\\
5 & 199.49 & 30 & 1.37 & 30\\
6 & 256.67 & 20 & 0.80 & 50\\
7 & 389.98 & -15 & 0.00 & 100
\end{tabular}
\caption{Ideal braking responses under several conditions and expected jump distances.}
\label{tabTests}
\end{table}

\begin{figure}[H]
\centering
\begin{tabular}{cc}
\includegraphics[width=0.45\textwidth]{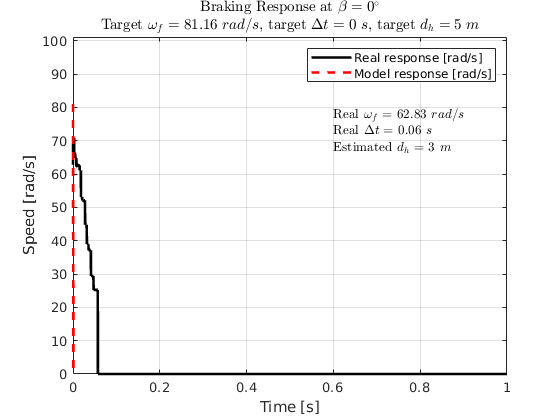} &
\includegraphics[width=0.45\textwidth]{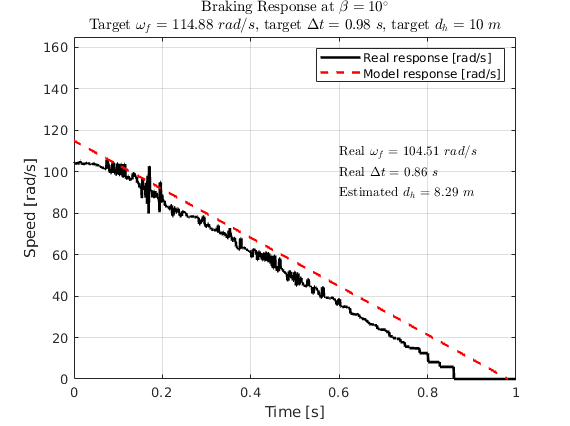}\\
\includegraphics[width=0.45\textwidth]{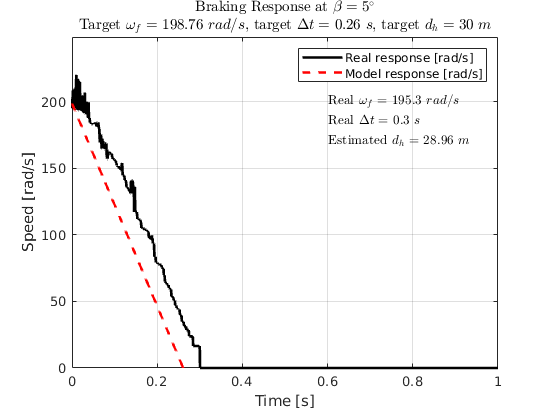} &
\includegraphics[width=0.45\textwidth]{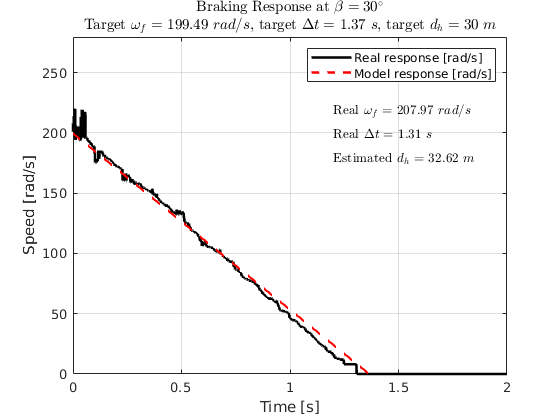}\\
\includegraphics[width=0.45\textwidth]{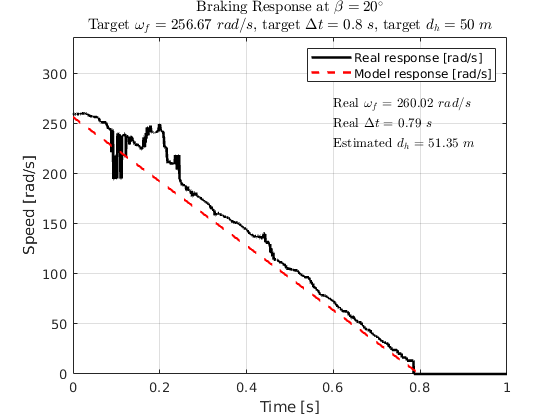} &
\includegraphics[width=0.45\textwidth]{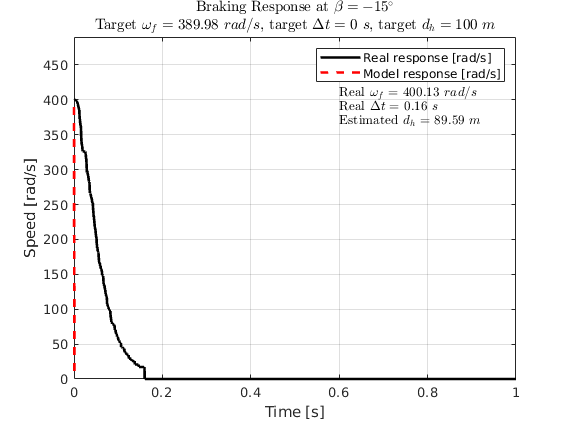}
\end{tabular}
\caption{Examples of the application of the braking by voltage inversion and the estimated distance that the robot is able to reach.}
\label{figSixExamples}
\end{figure}

The different plots of Figure \ref{figSixExamples} illustrate the responses of the controlled braking of the inertial wheel. Several inclinations and target distances have been selected to illustrate the behavior of the motor braking. The red lines indicate the ideal controlled speed of the motor and how it decreases from a designated $\omega_f$ to 0 in a $\Delta t$ time. Figure \ref{figSixExamples} shows the behavior of the tests proposed in Table \ref{tabTests}. Table \ref{tabSixExamples} makes a comparison of the obtained results. The relative error in percentage in Table \ref{tabSixExamples} is obtained from Equation \eqref{relError}

\begin{equation}
\label{relError}
Error = \frac{|expected\ response - real\ response|}{|expected\ response|} \times 100.
\end{equation}

\begin{table}[H]\centering
\begin{tabular}{ccrr|r}
\textbf{Slope} & \textbf{Parameter} & \textbf{Exp. Response} & \textbf{Real Response} & \textbf{Error}\\
\hline
0º & $d_h\ [m]$ & \textbf{5} & \textbf{3.00} & 40.00 \%\\
 & $\omega_f\ [rad/s]$ & 81.16 & 62.83\\
 & $\Delta t\ [s]$ & 0.00 & 0.06\\
\hline
10º & $d_h\ [m]$ & \textbf{10} & \textbf{8.29} & 17.10\%\\
 & $\omega_f\ [rad/s]$ & 114.88 & 104.51\\
 & $\Delta t\ [s]$ & 0.98 & 0.86\\
\hline
5º & $d_h\ [m]$ & \textbf{30} & \textbf{28.96} & 3.47\%\\
 & $\omega_f\ [rad/s]$ & 198.76 & 195.3\\
 & $\Delta t\ [s]$ & 0.26 & 0.30\\
\hline
30º & $d_h\ [m]$ & \textbf{30} & \textbf{32.62} & 8.73\%\\
 & $\omega_f\ [rad/s]$ & 199.49 & 207.97\\
 & $\Delta t\ [s]$ & 1.37 & 1.31\\
\hline
20º & $d_h\ [m]$ & \textbf{50} & \textbf{51.35} & 2.7\%\\
 & $\omega_f\ [rad/s]$ & 256.67 & 260.02\\
 & $\Delta t\ [s]$ & 0.80 & 0.79\\
\hline
-15º & $d_h\ [m]$ & \textbf{100} & \textbf{89.59}\\
 & $\omega_f\ [rad/s]$ & 389.98 & 400.13 & 10.41\%\\
 & $\Delta t\ [s]$ & 0.00 & 0.16
\end{tabular}
\caption{Comparison between the simulated and resultant responses of Figure \ref{figSixExamples}.}
\label{tabSixExamples}
\end{table}

\subsection{Jump Results}

For the simulations it has been selected the asteroid Itokawa with dimensions of $535 \times 294 \times 209$ m \cite{jpl2003itokawa}; also this asteroid has been selected because most of the estimated population of Near Earth Objects are under $1\ km$ of diameter \cite{harris2015population}. It is essential to remark that these specifications are merely speculative and the robot could be bigger and heavier depending on the on-board instruments and the tasks that the robot will perform.

To validate the resultant parabolic trajectories of the explorer, the surface slope was divided each 5º from 30º to -30º; for every angle, representative distances have been selected (5 $m$, 10 $m$, 30 $m$, 50 $m$, 70 $m$, and 100 $m$); for each of these distances, 14 repetitions of the braking of the DC motor and their consequent simulated jumps under the gravity of the asteroid Itokawa were conducted. Given a total of 1092 conducted experiments.

In this research, three surface inclinations have been chosen for the presentation of the results: 15º, 0º, and -15º. From these angles, have been plotted three target distances: 5 $m$, 50 $m$, and 100 $m$, in which the robot has reached, respectively, the worst, the best and an average performance according to the landing point and the original objective distance.

\begin{figure}[H]
\begin{tabular}{ccc}
\includegraphics[width=0.33\textwidth]{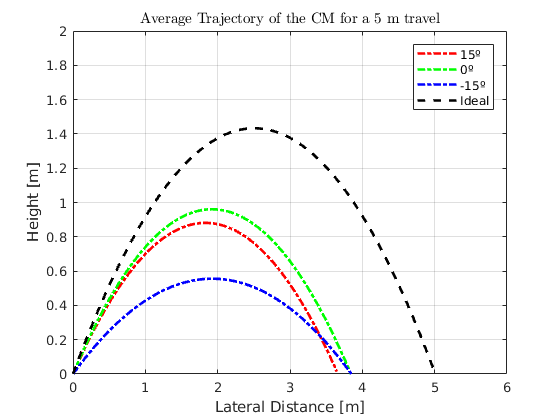} & \includegraphics[width=0.33\textwidth]{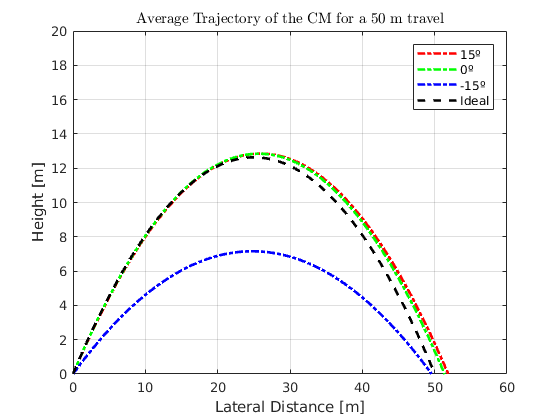} & \includegraphics[width=0.33\textwidth]{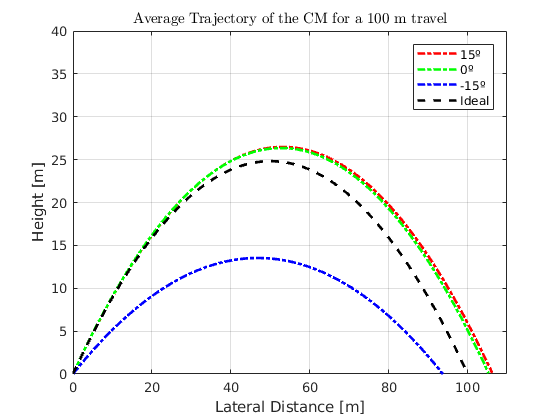}\\
\includegraphics[width=0.33\textwidth]{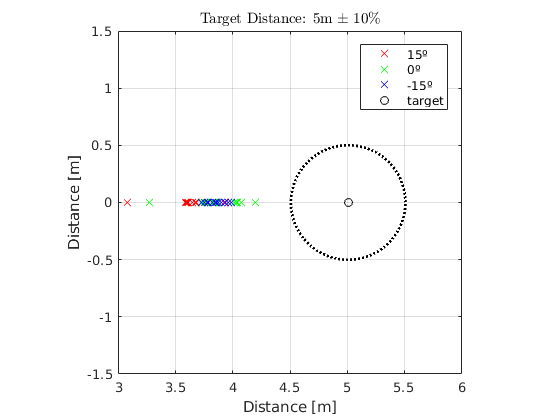} & \includegraphics[width=0.33\textwidth]{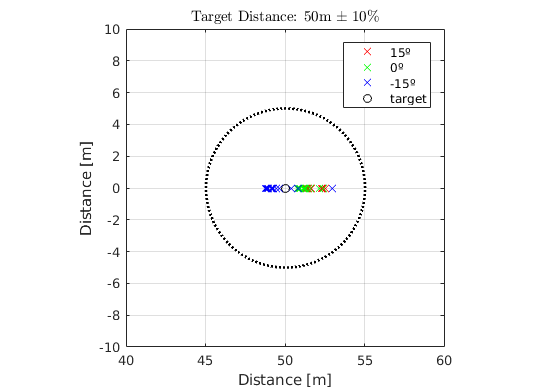} & \includegraphics[width=0.33\textwidth]{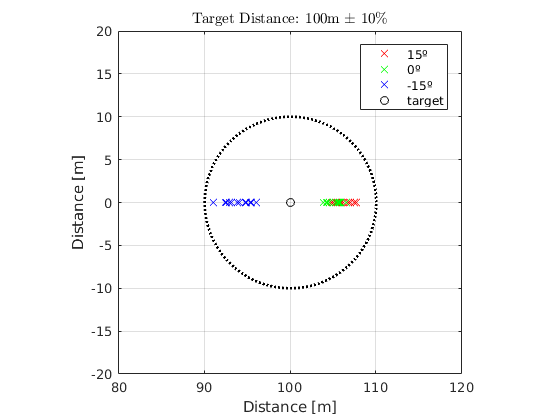}
\end{tabular}
\caption{Average trajectories and target distances.}
\label{average}
\end{figure}

In these tests, it has been conducted a series of 14 jump simulations for each of the three selected inclinations to reach 5 $m$, 50 $m$, and 100 $m$ targets. Every simulated jump has been obtained from a real braking experiment of the motor. It is expected that for every target distance, each landing will be within a 10\% tolerance area. In Figure \ref{average}, the first row of images shows the average parabolic trajectories for 5 m, 50 m, and 100 m travels, respectively, as well as the ideal flight. The second row offers a perspective from above of the target distance $d_h$ ($\circ$) with a 10\% error tolerance and the place where every simulation touched the ground ($\times$). As it can be seen, the worst result was scored in the 5m range with a relative error of 24.4\%, an average result was scored in the 100 m range with a relative error of 5.9\%, and the best result was scored in the 50 m range with a relative error of 2.4\%. Due to the inertia, it is physically impossible to stop the flywheel in $0$ s without impacting brakes, but as we have reviewed, braking a DC motor by voltage inversion is a good trade-off.

\subsection{Summary of the Results}\label{secSummary}

In this section, the summarized jump results described in the above subsection are presented.

\subsubsection{Surface Inclination $\beta = 15$º}

From Table \ref{tabRes15} it can be seen that the best response is in the 50 $m$ range with a relative error of 3.7\% and the worst is in the 5 $m$ distance with a relative error of 27.2\%.

\begin{table}[H]\centering
\begin{tabular}{crrr}
\textbf{Target Distance} & \textbf{Mean Distance} & \textbf{Std Deviation} & \textbf{Relative Error}\\
\hline
5 $m$ & 3.6 $m$ & $\pm0.2\ m$ & \textbf{27.2\%}\\
10 $m$ & 8.5 $m$ & $\pm0.2\ m$ & \textbf{15.5\%}\\
30 $m$ & 35.7 $m$ & $\pm1.6\ m$ & \textbf{19.1\%}\\
50 $m$ & 51.9 $m$ & $\pm0.4\ m$ & \textbf{3.7\%}\\
70 $m$ & 73.6 $m$ & $\pm0.2\ m$ & \textbf{5.1\%}\\
100 $m$ & 106.3 $m$ & $\pm0.9\ m$ & \textbf{6.2\%}
\end{tabular}
\caption{Summary of the results of $\beta = 15$º.}
\label{tabRes15}
\end{table}

\subsubsection{Surface Inclination $\beta = 0$º}

From Table \ref{tabRes0} it can be noticed that the best response is in the 50 $m$ range with a relative error of 2.7\% and the worst is in the 5 $m$ distance with a relative error of 23\%.

\begin{table}[H]\centering
\begin{tabular}{crrr}
\textbf{Target Distance} & \textbf{Mean Distance} & \textbf{Std Deviation} & \textbf{Relative Error}\\
\hline
5 $m$ & 3.8 $m$ & $\pm0.2\ m$ & \textbf{23.0\%}\\
10 $m$ & 8.5 $m$ & $\pm0.2\ m$ & \textbf{14.8\%}\\
30 $m$ & 31.4 $m$ & $\pm1.9\ m$ & \textbf{4.8\%}\\
50 $m$ & 51.4 $m$ & $\pm0.5\ m$ & \textbf{2.7\%}\\
70 $m$ & 73.0 $m$ & $\pm0.4\ m$ & \textbf{4.2\%}\\
100 $m$ & 105.3 $m$ & $\pm1.0\ m$ & \textbf{5.3\%}
\end{tabular}
\caption{Summary of the results of $\beta = 0$º.}
\label{tabRes0}
\end{table}

\subsubsection{Surface Inclination $\beta = -15$º}

From Table \ref{tabRes-15} it can be noticed that the best response is in the 50 $m$ range with a relative error of 0.8\% and the worst is in the 5 $m$ distance with a relative error of 23.1\%.

\begin{table}[H]\centering
\begin{tabular}{crrr}
\textbf{Target Distance} & \textbf{Mean Distance} & \textbf{Std Deviation} & \textbf{Relative Error}\\
\hline
5 $m$ & 3.9 $m$ & $\pm0.1\ m$ & \textbf{23.1\%}\\
10 $m$ & 8.4 $m$ & $\pm0.3\ m$ & \textbf{16.1\%}\\
30 $m$ & 34.1 $m$ & $\pm2.2\ m$ & \textbf{13.7\%}\\
50 $m$ & 49.6 $m$ & $\pm1.2\ m$ & \textbf{0.8\%}\\
70 $m$ & 66 $m$ & $\pm0.7\ m$ & \textbf{5.7\%}\\
100 $m$ & 93.8 $m$ & $\pm1.4\ m$ & \textbf{6.2\%}
\end{tabular}
\caption{Summary of the results of $\beta = -15$º.}
\label{tabRes-15}
\end{table}

\subsection{Consecutive Jumps. Big distances: 385 m}

\begin{figure}[H]\centering
\begin{tabular}{cc}
\includegraphics[width=0.45\textwidth]{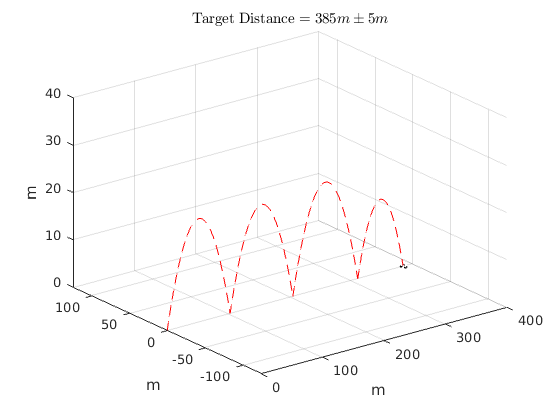}&
\includegraphics[width=0.45\textwidth]{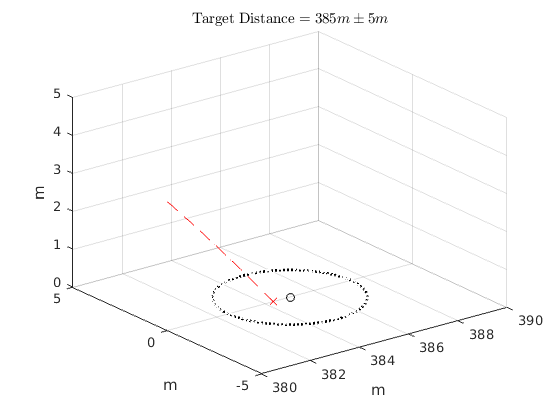}
\end{tabular}
\caption{Demonstration of an ideal path to reach the objective target of 385 $m$ with a 5 $m$ of tolerance over flat terrain.}
\label{fig385}
\end{figure}

As it has been said in Section \ref{deflecting}, distances greater than 100 $m$ are possible, but reaching the escape velocity is a considerable risk to the explorer. To prevent this, it might be preferable that the cubic robot conducts a series of jumps to go to the desired place. For a $d_h = 385\ m$ with 5 $m$ error tolerance, three jumps of 100 $m$ and one jump of 85 $m$ are needed. It is expected that the \textit{hopper} lands within the tolerance area. In Figure \ref{fig385} the trajectories of the four jumps needed for a target distance of 385 $m$ over flat terrain are shown. In the same way, as in the previous section, the parabolic flights are simulated from the braking responses of the flywheel under the gravitational pull of the asteroid Itokawa. Table \ref{tab385} presents the results of these jumps.

\begin{table}[H]\centering
\begin{tabular}{cr|r}
\textbf{Target Distance $[m]$} & \textbf{Resultant Distance $[m]$} & \textbf{Error} [\%]\\
\hline
100 & 102.41 & 2.41\\
100 & 102.18 & 2.18\\
100 & 105.67 & 5.67\\
75 & 73.95 & 1.40\\
\textbf{385} & \textbf{384.33} & 0.17
\end{tabular}
\caption{Simulated jumps of the \textit{hopper}. The accumulation of the resultant distances reduced the length of the last jump.}
\label{tab385}
\end{table}

\section{Conclusions and Future Work}

A new kind of space robots with locomotion based on jumps developed in this research provides a new approach for the exploration of small bodies of our Solar System, taking advantage of the low gravity of these interplanetary objects.

We proposed a new way of stopping the flywheel no suddenly but in a short lapse by means of voltage inversion.

INAHOPPER has a cubic small shape with no external actuators. The necessary momentum for executing a jump is provided by internal inertial wheels. This approach comprises three basic steps: 1) develop the mathematical model of the dynamics of the robot; 2) implement a speed controller for the inertial wheels in a prototype of the robot; and 3) with the results of the experiments in the prototype simulate the trajectory of the robot over some asteroid of interest.

A prototype of the rover was built to verify the efficiency of the proposed speed controller of the flywheel. Computer simulations that emulate the parabolic jumps of the rover under milli-microgravity environments according to the real responses of the braking of the inertial wheel were developed.

Considering the examples given in Section \ref{secResults} it is possible to observe the behavior of the braking responses. Due to the inertia of the flywheel used in the prototype of this research, stopping angular speeds higher than $350\ rad/s$ in a short braking time is difficult to achieve, but it is still a good trade-off. The response of the DC motor achieved the expected behavior provided that the angular speeds is in between of $150\ rad/s$ and $350\ rad/s$. For angular speeds lower than $150\ rad/s$, the objective angular speed is not reached for the DC motor and its response does not fit the expected results causing the biggest relative errors. The jump distance results from the response of the DC motor. Section \ref{secSummary} indicates that large courses $d_h > 30\ m$ have less than 10\% of error, but lower distances have greater errors; however, this approach of locomotion has the advantage that if the robot does not reach the indicated place in the first jump, it can perform a series of jumps to reach a given location.

In the near future, we plan to extend the analysis to the 3 degrees of freedom of INAHOPPER. Traditionally the DC motors are considered as linear time-invariant models, but in practice, controlling high and low velocities could be a nonlinear problem, for this reason, develop a system capable of controlling nonlinear systems is important. Also, it will be essential to study the bounce over different kind of soils and consider irregular gravity fields.

{\footnotesize \textbf{Author Contributions:} All authors made substantial contribution to this research. A. Batrez designed and implemented the original control algorithm, and conducted simulations and experiments to evaluate its performance, and wrote the first journal manuscript. G. Mendoza revised the control model. G. Rodriguez and A. Muñoz revised algorithms and models and supervised the research. All authors discussed and interpreted the results, and, agreed about the conclusions.}

{\footnotesize \textbf{Funding:} The project INAHOPPER was supported by the Mexican National Council for Science and Technology, Grant No. 776915.}

\appendix
\section{Appendix}
\unskip
\subsection{}\label{app1}

The surface inclination $\beta$ will be positive for counterclockwise angles and negative for clockwise angles.

For the leverage phase, the equations of motion are based on the inverted pendulum model, Equation \eqref{hop1} and \eqref{hop2}, and are based in \cite{ogata2010ingenieria}
\begin{equation}\label{hop1}
\left(I + m l^2 \right) \ddot{\theta} + m l \ddot{x} = m g l \theta,
\end{equation}
equating $m l \ddot{x} = \tau$ and the inertia $I = I_p$, we have an expression for the angular velocity of the pendulum $\ddot{\theta}$
\begin{equation}\label{hop2}
\ddot{\theta}  = \frac{m_p g l \sin \theta - \tau}{I_p + m_p l^2}.
\end{equation}

\citet{allen2013internally} have derived in Equation \eqref{hop3} that the minimum torque to initiate a rotation from rest is
\begin{equation}\label{hop3}
\tau_{min} = m_p g l \sin (\alpha + \beta),
\end{equation}
also, the cited author have defined that the energy transfer ratio as indicated in Equation \eqref{hop4}
\begin{equation}\label{hop4}
\eta = \frac{E^-}{E^+} = \frac{I_f}{I_p + m_p l^2},
\end{equation}
where $E^-$ is the energy just before actuation (flywheel kinetic energy) and $E^+$ is the energy just after actuation (platform kinetic energy) \cite{hockman2017design}.

By combining the angular momentum of the flywheel to the angular momentum of the platform about a spike we obtain Equation \eqref{hop5}
\begin{equation}\label{hop5}
I_f \omega_f = \dot{\theta} ( I_p + m_p l^2 ),
\end{equation}
and by substituting Equation \eqref{hop5} in Equation \eqref{hop4} it is obtained the resulting hop velocity $v_h$. Assuming an immediate momentum transfer, $v_h = l \dot{\theta}$, the hop velocity is obtained by Equation \eqref{hop7}
\begin{equation}\label{hop7}
v_h = \frac{I_f \omega_f l}{I_p + m_p l^2}.
\end{equation}

\bibliographystyle{IEEEtranN}
\bibliography{references.bib}

\end{document}